\def\year{2022}\relax
\documentclass[letterpaper]{article} % DO NOT CHANGE THIS
\usepackage{aaai22}  % DO NOT CHANGE THIS
\usepackage{times}  % DO NOT CHANGE THIS
\usepackage{helvet}  % DO NOT CHANGE THIS
\usepackage{courier}  % DO NOT CHANGE THIS
\usepackage[hyphens]{url}  % DO NOT CHANGE THIS
\usepackage{graphicx} % DO NOT CHANGE THIS
\def\UrlFont{\rm}  % DO NOT CHANGE THIS
\usepackage{natbib}  % DO NOT CHANGE THIS AND DO NOT ADD ANY OPTIONS TO IT
\usepackage{caption} % DO NOT CHANGE THIS AND DO NOT ADD ANY OPTIONS TO IT
\DeclareCaptionStyle{ruled}{labelfont=normalfont,labelsep=colon,strut=off} % DO NOT CHANGE THIS
\usepackage{algorithm}
\usepackage{algorithmic}
\usepackage{xcolor}
\usepackage{booktabs}
\usepackage{amsfonts,amssymb,amsmath}
\DeclareMathOperator*{\argmax}{arg\,max}

\usepackage{newfloat}
\usepackage{listings}
\floatstyle{ruled}
\newfloat{listing}{tb}{lst}{}
\floatname{listing}{Listing}
\title{Grounding Object Relations in Language-Conditioned Robotic Manipulation \\with Semantic-Spatial Reasoning}
\author{
    Qian Luo\textsuperscript{\rm 1*},% \thanks{XXXX.}
    Yunfei Li\textsuperscript{\rm 2*},
    Yi Wu\textsuperscript{\rm 1,2}
}
\affiliations{
    *Indicates equal contribution\\
    \textsuperscript{\rm 1}Shanghai Qi Zhi Institute\\
    \textsuperscript{\rm 2}Institute for Interdisciplinary Information Sciences, Tsinghua University\\
    luoqian19961224@gmail.com, liyf20@mails.tsinghua.edu.cn, jxwuyi@gmail.com
}

\usepackage{bibentry}
\begin{document}
\setcounter{secnumdepth}{2}

\maketitle

\begin{abstract}
Grounded understanding of natural language in physical scenes can greatly benefit robots that follow human instructions. In object manipulation scenarios, existing end-to-end models are proficient at understanding semantic concepts, but typically cannot handle complex instructions involving spatial relations among multiple objects. which require both reasoning object-level spatial relations and learning precise pixel-level manipulation affordances. % , which hinders their application to complex scenes xxx. 
We take an initial step to this challenge with a decoupled two-stage solution. In the first stage, we propose an object-centric semantic-spatial reasoner to select which objects are relevant for the language instructed task. The segmentation of selected objects are then fused as additional input to the affordance learning stage. Simply incorporating the inductive bias of relevant objects to a vision-language affordance learning agent can effectively boost its performance in a custom testbed designed for object manipulation with spatial-related language instructions.
\end{abstract}

\section{Introduction}
Understanding complex language instructions is a long-standing research problem for building intelligent robots that can assist humans to perform meaningful tasks in grounded scenes~\cite{chen11learning,bollini12interpreting,misra16tell-me-dave}. Recently, there has been exciting progress in grounded vision and language in robotic manipulation scenarios with reinforcement learning and imitation learning~\cite{nair21learning,jang21bc-z}. Leveraging the power of pretrained vision and language models, some most advanced end-to-end models can effectively ground semantic concepts from natural language to physical scenes and even demonstrate some generalization ability to unseen objects~\cite{shridhar2022cliport}. 

\begin{figure*}
    \centering
    \includegraphics[width=1.0\linewidth]{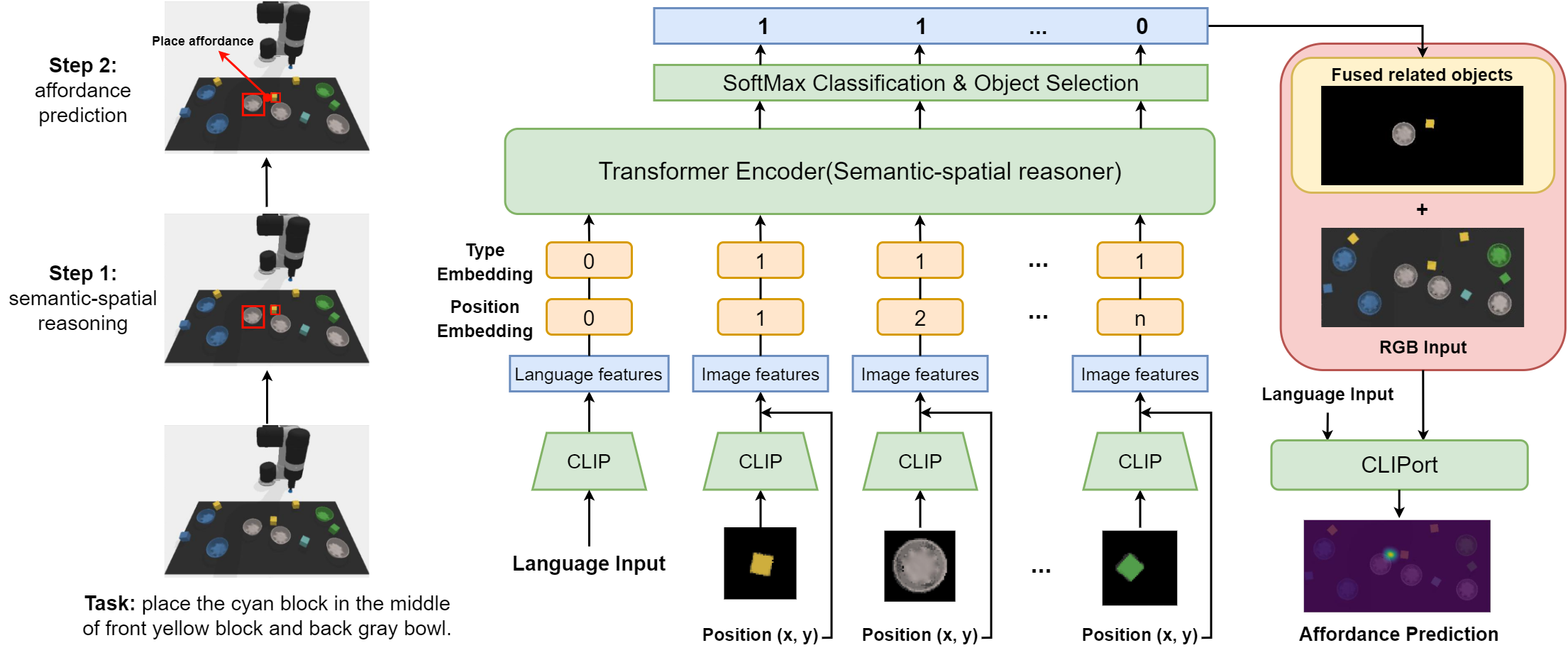}
    \caption{Overview of our two-stage framework for grounding spatial-related instructions in object manipulation tasks.
    }
    \label{fig:overview}
\end{figure*}

However, existing end-to-end agents in vision-language manipulation typically still lack the ability to deal with instructions containing \textit{spatial relations among multiple objects}. Consider a motivating example in Fig.~\ref{fig:overview} where multiple blocks and bowls with identical or different colors are put on a table. Existing approaches can already follow instructions such as ``\textit{put the cyan block into a green bowl}'' but cannot handle a more complex instruction like ``\textit{place the cyan block in the middle of the front yellow block and the back gray bowl}''. Solving the previous task only requires understanding simple semantic concepts while the latter one additionally requires the agent to reason over more fine-grained spatial relations.    % \yf{ However, the existing methods typically focus on problems that only require simple entity concept understanding (e.g., ``put the red block to the green bowl''), and they lack the grounded understanding of more complex \textit{spatial relations} about multiple objects, e.g., they are poor at following the instruction as ``put to the middle of A and B''. }

We hypothesize that current end-to-end models struggle at such complex instructions since it is challenging to simultaneously learn both abstract spatial reasoning and precise manipulation skills in a fully coupled manner.
In this paper, we take an initial step towards this challenging problem by decoupling object-level spatial reasoning and pixel-level affordance learning in two stages. We propose an object-centric semantic-spatial reasoning module that predicts which objects are relevant for accomplishing a manipulation task from the language instruction and image observation. Then we fuse the segmentation of all the relevant objects to the input of a strong performing vision-language affordance learning agent CLIPort~\cite{shridhar2022cliport}. The relevant object prediction serves as an effective inductive bias to inform the agent about which regions are beneficial to look at for low-level affordance learning. 

To better evaluate whether our proposed framework can effectively handle spatial relations between objects, we design a custom pick-and-place task in cluttered scenes built upon Ravens~\cite{zeng2021transporter}. The language instructions are designed to involve relative spatial positions of objects. Preliminary results show that our method achieves non-trivial success in this task while CLIPort simply cannot work given the same amount of training data. % \yf{any more to add?}

\section{Related Work}
\textbf{Language-conditioned robotic manipulation:}
% Earlier work in language-condition manipulation
Instructing robots with natural language has attracted much research interest in recent years~\cite{mees22calvin}. 
Some earlier works proposed end-to-end learning methods for vision-language-conditioned control by imitating large-scale behavior datasets~\cite{lynch21language,jang21bc-z,mees22what} or learning reward functions for reinforcement learning~\cite{shao20concept2robot,nair21learning}. % (Imitation Learning, RL, ...)
With advances in multi-modal representation learning~\cite{radford2021learning}, there are also many works leveraging pretrained vision or language encoders for grounded manipulation tasks~\cite{shridhar2022cliport,mohit22perceiver-actor}, and exciting progress in utilizing the planning ability of large language models for embodied reasoning tasks~\cite{saycan,socratic-models,innermonologue}.
% Recent Advances. (SayCan...)
We are particularly interested in grounding language instructions with complex spatial relations between objects, which cannot be addressed well by previous methods.
% Limitation of these work: without spatial-reasoning, can only deal with semantic-related cases.

\textbf{Semantic-spatial reasoning for manipulation:}
% Neuro-Symbolic, StructFormer, SORNet, ...
Learning spatial relations between different objects have been studied in vision and robotics~\cite{johnson17clevr,mees17metric}, which can better address language goals specifying the desired spatial relations~\cite{paul16efficient,venkatesh21spatial,yuan21sornet,liu22structformer}, and can be benefit hierarchical planning to accomplish complex goal configurations~\cite{zhu21hierarchical}. 
% Semantic-spatial reasoning~\cite{johnson17clevr,sharma20relational} has been studied recent years to deal with complex language-guided manipulation tasks where the spatial relationships between objects are involved. 
A notable line of works study neural-symbolic reasoning to model spatial relations and semantic concepts in a scene~\cite{mao19neuro-symbolic}. More recently, fully end-to-end learning mechanism has also achieved much progress in modeling complex object-level relations for visual reasoning with correct inductive bias~\cite{ding21attention}. 
In this work, we also propose a purely data-driven model for semantic-spatial reasoning which is easy to implement and can improve with the increasing scale of the training dataset.
% The understanding of semantic information and spatial relationships in a scene can be efficiently grasped by neural-symbolic reasoning. ~\cite{mao19neuro-symbolic}

% modeling spatial relations in vision and robotics: ~\cite{mees17metric,sharma20relational,}
%  neural-symbolic~\cite{mao19neuro-symbolic}, fully learned~\cite{ding21attention};
% ground spatial relations in natural language to robot manipulation~\cite{paul16efficient,venkatesh21spatial,liu22structformer}: ;  
% \yf{Pretraining in grounded visual-language undertanding} (not very related)

\section{Method}
We propose a two-stage solution for grounding instructions with spatial relations between objects. In the first stage, a semantic-spatial reasoning module aims to model the object-level relations from language instructions and object-centric image observations (see Sec.~\ref{sec:method:semantic-spatial}). % that can effectively incorporate object-level inductive bias into cliport for better grounding instructions with complex object relations. 
% The semantic-spatial module takes both \textcolor{blue}{Clip processed} language instructions and object-centric image patches \textcolor{blue}{and positions} as input to select related objects to the tasks. 
In the second stage, the segmentation of the selected objects is fed as an additional input to a pixel-level affordance learning model CLIPort, and functions as an inductive bias that guides the agent to attend to particular objects when predicting precise pick and place affordance maps (see Sec.~\ref{sec:method:affordance-prediction}). Combining with an off-the-shelf object detection and segmentation model, the two stages are cascaded together in deployment (Sec.~\ref{sec:method:pipeline}).

% \begin{figure}
%     \centering
%     \includegraphics[width=1.0\linewidth]{figures/model.png}
%     \caption{Overview
%     }
%     \label{fig:model}
% \end{figure}

\subsection{Object-centric semantic-spatial reasoning}\label{sec:method:semantic-spatial}
% \yf{definitions of input and output; transformer architecture; how do we get labels; the objective of this module}
The semantic-spatial reasoning (SSR) model aims to select relevant objects in a scene for the pick-and-place task %\textcolor{red}{placement?} \textcolor{blue}{pick and placement, in the experiment we only use placement} 
given a complex language instruction containing semantic and spatial relations of objects. We leverage a powerful Transformer~\cite{vaswani2017attention} architecture to allow grounded understanding of relations from natural language instructions and orthographic RGB images. 
Specifically, the module takes in an instruction $(w_1, w_2, \cdots, w_L)$, and $m$ image patches with shape $50\times 50 \times 3$ center-cropped from the detected positions of all the objects along with their $(u, v)$ coordinates in the orthographic image. $L$ is the length of the language instruction and $m$ is the total number of objects in the image. 
The words and image patches are encoded with the pretrained CLIP model~\cite{radford2021learning}. The coordinates of objects are encoded with linear layers, then concatenated with the corresponding image patch embeddings into $m$ object-centric features. The $L$ word features and $m$ object-centric features are then concatenated with their position embeddings. To distinguish text and object embeddings, we additionally concatenate type embeddings (0 for texts and 1 for objects) to each token. The $L+m$ tokens are then passed into 8 self-attention layers. The fused features at $m$ object-centric tokens are used to predict scores $\hat{s}_i$ of how related each object is to the language instruction with an MLP. The ground truth for this reasoning module is represented as a binary vector $[s_1, s_2, \cdots, s_m]$, where $s_j=1$ if the $j^{th}$ object is related for performing placement and $s_j=0$ if the object is irrelevant.
Finally, we normalize the scores and the ground truth with softmax and optimize the module to minimize the $L_1$ distance between them.  

\textbf{Implementations}: % \yf{how do we get object-centric image patches and bounding boxes;} 
% \textcolor{blue}{In Ravens, we can get one-hot mask of every object in a top down map, and can also get the rgb mask by multiplying rgb top down map. We propose that the object size(pixel area) not exceed the crop size(50, 50). The object will appear in the middle of object-centric image. If the object is too close to the boundary, we will force the cropped image appear inside the boundary.} 
Our semantic-spatial reasoning module requires cropped object-centric patches as input, which can be predicted from off-the-shelf object detection models. In practice, we adopt an open-vocabulary object detection model ViLD~\cite{gu2021open} that is suitable for detecting arbitrary categories of objects to predict the locations of all objects. A cheaper way to obtain the image patches when training in simulation is to directly project the center of each object onto the orthographic image, then crop patches with shape $50\times 50\times 3$ centered at the projected coordinates. % We also experiment with oracle coordinates of objects projected to the orthographic images from the ground-truth states in simulation, which is cheaper to obtain. The ground-truth related objects are obtained using previledged information when generating task instances in simulation.

\subsection{Affordance prediction with selected objects}\label{sec:method:affordance-prediction}
In this part, we describe how we leverage object-level relevance to help low-level manipulation learning.

We adopt a similar two-stream network architecture as CLIPort for predicting pick and place affordance, where a semantic pathway built upon a pretrained CLIP encoder and a spatial pathway based on a transporter network are fused together to predict the pixelwise probability mass $\mathcal{Q}_{\textrm{pick}, \textrm{place}} \in \mathbb{R}^{H\times W}$. For more details, we encourage the readers to refer to the CLIPort paper~\cite{shridhar2022cliport}.

The only difference in our architecture is that our spatial pathway additionally takes as input an attention map indicating which objects may be relevant to the manipulation task. In this way, the affordance prediction model is informed of the high-level semantic-spatial reasoning result and can focus more on the learning of low-level affordance. 
% \yf{how we get the related segmentation; }

The attention map is calculated as follows.
After predicting the scores of how relative different objects are to the manipulation task, we select the objects with normalized scores greater than a threshold to guide the affordance prediction stage. We combine the binary segmentation mask of all the selected objects (by computing their logical ``OR'') and multiply the mask by the original RGB image to get an attention map with the same shape $(H, W, 3)$ as the original image. 
The attention map is stacked in channel dimension with the original RGB image as the input to the spatial pathway. 

The ground truth of the affordance prediction model is the projected pick and place poses to the 2D image from demonstrations. The training objective is the cross-entropy between normalized $\mathcal{Q}_{\textrm{pick},\textrm{place}}$ and the ground truth.

\subsection{Training and deployment}\label{sec:method:pipeline}
The semantic-spatial reasoning module and the affordance prediction module are trained separately with supervised learning. We experiment with using oracle object detection and also ViLD detection as input to train the semantic-spatial reasoning module. For affordance prediction module, we always use ground-truth relevant object segmentation as its input during training. In practice, we find our semantic-spatial reasoning module is lightweight for training and can benefit from training over large-scale datasets while training the affordance prediction module is time-consuming and we can only afford training on 1k demos with our best efforts. % \yf{decoupled training of the light-weight reasoning module and the heavy cliport module. Mention that with decoupled training, one can use large-scale datasets to achieve strong performance while not sacrificing much training time.}

During deployment, we first compute the coordinates, cropped patches and segmentation masks of all the objects by querying a ViLD model, then feed the object-centric information along with the language instruction into  the semantic-spatial reasoning module to get normalized relevance scores for all the objects. The objects with scores greater than the average are selected and fused into the affordance prediction model to get $\mathcal{Q}_{\textrm{pick}}$ and $\mathcal{Q}_{\textrm{place}}$. The best pick/place location is then computed as $\underset{(u, v)}\argmax \mathcal{Q}(u, v)$. Using the parameters of the camera from factory calibration and hand-eye calibration, the 2D locations on the orthographic image can be easily converted to 3D positions relative to the robot base frame. The robot can run any suitable motion planner to reach the desired pick/place poses. 
% \textbf{Implementations}
%  how the two stages are trained (and fine-tuned together?)

\section{Experiments}
In this section, we discuss the simulation experiments for validating the proposed method.
We design the experiments to answer the following questions.
\begin{enumerate}
    \item Can our method effectively solve a typical spatial-related pick and place task in Ravens environment?
    \item How well does our method perform compared with the baseline method and those validated via ground truth?
\end{enumerate}

\subsection{Experimental setup}\label{sec:expr:experimental-setup}
\subsubsection{Task setup}
% \textbf{A new task for grounded object relation understanding}

% \textbf{existing benchmark tasks?}

We build our task upon Ravens, a collection of simulated tasks in PyBullet~\cite{coumans2020} for learning vision-based robotic manipulation. 
We study a custom language-conditioned pick-and-place task that requires semantic-spatial reasoning.
At the beginning of each episode, 6 bowls and 6 blocks are generated at random positions on the table, and objects within each category are of identical shape. 
Three blocks are of color A, three bowls are of color B and the other six objects are of random colors different from A and B. % of them share the same given color while the others have 3 different colors from the given one. 
% Thus, we have 3 identical blocks (of color A) and bowls (of color B) and other 6 objects(color X bowl/block, X are different from each other), 12 in total.
To test the capability of spatial understanding of our model, the model is supposed to identify the spatial relations among those identical blocks and bowls. 
The language instructions are generated from the template ``pick the \texttt{[color X]} block in the middle of \texttt{[location a]} \texttt{[color A]} block and \texttt{[location b] [color B]} bowl'', in which the locations are sampled from left/right/front/back.

\begin{table*}
\centering
\caption{\label{tab:results1}Pick-and-place success rate of different agents evaluated with seen and unseen objects.}   
\begin{tabular}{lllllllll}     
\toprule & \multicolumn{2}{c}{100 demos} & \multicolumn{2}{c}{1000 demos} & \multicolumn{2}{c}{10k demos} & \multicolumn{2}{c}{100k demos}\\
   & Seen & Unseen & Seen & Unseen & Seen & Unseen & Seen & Unseen\\    \midrule  
 Cliport & 0 & 0 & 0 & 0 & N/A & N/A & N/A & N/A\\
 Cliport oracle & 0.45 & 0.20 & 0.90 & 0.59 & N/A & N/A & N/A & N/A \\
 SSR + all mask & 0.08 & 0.02 & 0.36 & 0.14 & 0.68 & 0.34 & 0.71 & 0.35 \\
 SSR + ViLD & 0.02 & 0.01 & 0.30 & 0.11 & 0.60 & 0.25 & 0.65 & 0.29 \\
 \bottomrule   
\end{tabular}  
\label{table:1}
\end{table*}

% \yf{Add a figure here illustrating the task setting}
\begin{figure}
    \centering
    \includegraphics[width=1.0\linewidth]{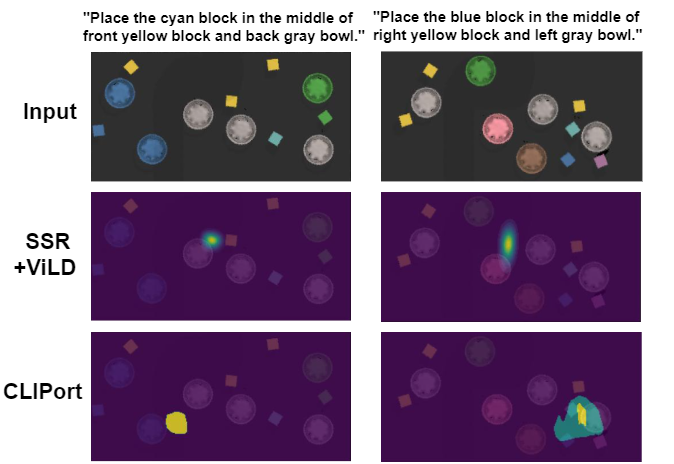}
    \caption{Predicted affordance maps for placement with different methods. Our method can accurately focus on the desired position while CLIPort completely fails. % Place affordance prediction
    }
    \label{fig:afford}
\end{figure}

\subsubsection{Training details}
Following the description in the task setup, we generate training, validation and test datasets in Ravens environment. 
Specifically, for each sample, the objects are added in accordance with the language instruction described in Sec.~\ref{sec:expr:experimental-setup}, where \texttt{location a} and \texttt{location b} are randomly chosen from left/right/front/back and \texttt{color X}, \texttt{color A}, \texttt{color B} are randomly selected from a collection of 7 colors.
The orthographic RGB image, the cropped oracle object-centric patches together with its projected ground-truth coordinates are added into training instances. 
The labels include the binary vectors representing the relevance of each object to the given language instruction, and the final pick and place coordinates for grounding.

The training procedures include the semantic-spatial reasoning stage and the affordance prediction stage. 
In the semantic-spatial reasoning stage, we feed the language instructions, oracle object-centric patches and position information into the transformer, and train the model to fit the normalized binary label. 
In the affordance prediction stage, we train the downstream CLIPort with segmentation fused from the ground truth of relevant objects. % \yf{oracle or predicted objects or both, be more concrete}.
% \textcolor{blue}{ground-truth relevant object}

% \yf{\textbf{Evaluation metrics}:}

% \begin{table*}
%     \centering
%     \caption{Pick and Place Results of Unseen Objects \textcolor{red}{TODO: Update numbers}}\label{tab:results2}
%     \begin{tabular}{lllllllll}  
%         \toprule & \multicolumn{2}{c}{100 demos} & \multicolumn{2}{c}{1000 demos} & \multicolumn{2}{c}{10k demos} & \multicolumn{2}{c}{100k demos}\\
%         & Success & Avg Dist & Success & Avg Dist & Success & Avg Dist & Success & Avg Dist\\    \midrule  
%         Cliport & 0 & 155 & 0 & 0 & 0 & 0 & 0 & 0\\
%         Selected mask & 0.96 & 22 & 0 & 0 & 0 & 0 & 0 & 0 \\
%         SSR + all mask & 0.82 & 42 & 0 & 0 & 0 & 0 & 0 & 0 \\
%         SSR + ViLD & 0.74 & 50 & 0 & 0 & 0 & 0 & 0 & 0 \\
%         \bottomrule   
%     \end{tabular}  
% \end{table*}

\subsection{Main results}
We train each agent with four datasets with different number of demonstrations (100, 1000, 10k, 100k) and choose the best checkpoints on the validation dataset (100 episodes) for evaluation. Each agent is evaluated with the average success rate on the test dataset consisting of 100 cases. % \yf{describe success criterion in task setup} 
An episode is regarded as successful if the correct object is picked and placed within 0.1m distance error to the target position. To test the generalization ability, we evaluate the agents in two settings where the object colors are seen or unseen during training.
We compare the performance of four agents:
\begin{itemize}
    \item \textbf{CLIPort:} The agent is trained and evaluated with CLIPort, an end-to-end model that directly predicts pick-and-place affordances from texts and RGBD images.
    \item \textbf{CLIPort with oracle relevant objects:} The agent is trained and evaluated by feeding CLIPort with ground-truth relevant objects. The purpose of this experiment is to verify whether fusing the segmented RGB image of relevant objects to CLIPort can enhance its ability to ground spatial-related language instructions. It approximates the upper bound of the performance if we have a perfect semantic-spatial reasoning module.
    \item \textbf{SSR + all mask:} The agent is trained on SSR module with ground-truth segmentation of all objects. The affordance prediction network is trained separately following the same procedure in \textbf{CLIPort with oracle relevant objects}. The agent is evaluated using ground-truth segmentation of all objects, and the SSR output is fed into the affordance prediction network. The purpose of this experiment is to test the ability of the semantic-spatial reasoning module.
    \item \textbf{SSR + ViLD:} The agent is trained following the same procedure in \textbf{SSR + all mask}. The only difference is that we use ViLD segmentation results instead of ground-truth segmentation of all objects. The agent is evaluated in an end-to-end manner without requiring any ground-truth information. The experiment results show how well our model performs without any oracle inputs.
\end{itemize}
The quantitative experiment results are summarised in Table \ref{table:1}. Since CLIPort already takes 2 days to train with 1000 demos, we cannot afford to train it with 10k and 100k demos. The \textbf{CLIPort oracle} agent is also trained on 100/1000 demos, which are used for the affordance prediction of \textbf{SSR + all mask} and \textbf{SSR + ViLD} agents. We also visualize predicted affordances in Fig.~\ref{fig:afford}.

Our methods (\textbf{SSR + ViLD} and \textbf{SSR + all mask}) outperform CLIPort in 100 and 1000 demos datasets. 
With larger datasets, all the agents gain better performance. In all cases, \textbf{SSR + all mask} performs slightly better than \textbf{SSR + ViLD} agent. The performance gap can be explained by imperfect predictions of ViLD. Our best agent is the \textbf{SSR + all mask} agent trained with 100k demos, which reaches a success rate of over 0.7. Since \textbf{CLIPort oracle} succeeds in 90 percent of the cases, there is still room to improve for SSR. Our agents are also able to generalize to unseen cases, but the performances are at most 50 percent as good as those in seen cases. We leave a more generalizable agent for future work.

\section{Conclusion}
In this work, we study the problem of grounded understanding of complex language instructions involving spatial relations between multiple objects in language-conditioned robotic manipulation. We take a preliminary step by enhancing the CLIPort with relevant object masks predicted from an object-centric semantic-spatial reasoning module. Experiment results show that informing the agent of which objects should be focused on is a simple yet effective way to address language-conditioned tasks that require understanding semantic-spatial relations. We plan to extend this work to more general cases with a greater variety of objects and manipulation tasks. Another direction is to work beyond pairwise relations (e.g., left/right) and to tackle multi-object relations (e.g., middle, pyramid).  

\bibliography{aaai22.bib}
\appendix
\end{document}